\documentclass[conference,a4paper]{APSIPA2026}
\usepackage{amsmath,amssymb,amsfonts}
\usepackage{graphicx}
\usepackage{multirow}
\usepackage{threeparttable}
\usepackage{booktabs}
\usepackage[backend=biber,style=ieee]{biblatex}
\usepackage{pgfplots}
\usepackage{geometry}
\usepackage{fancyhdr}

\fancypagestyle{firststyle}{
  \fancyhf{}
  \fancyhead[C]{2026 Asia Pacific Signal and Information Processing Association Annual Summit and Conference (APSIPA ASC)}
}

\begin{document}

\title{DeepFreqMark: End-To-End Learnable Frequency-Domain Watermarking with Spherical Attack Simulation for Latent Diffusion Models}

\author{
\authorblockN{
Chen-Hsiu Huang\authorrefmark{1}, 
Mario Köppen\authorrefmark{2} and
Ja-Ling Wu\authorrefmark{1}
}

\authorblockA{
\authorrefmark{1}
National Taiwan University, Taipei, Taiwan \\
E-mail: \{chenhsiu48,wjl\}@cmlab.csie.ntu.edu.tw}

\authorblockA{
\authorrefmark{2}
Kyushu Institute of Technology, Fukuoka, Japan \\
E-mail: mkoeppen@ieee.org}
}

\maketitle
\thispagestyle{firststyle}
\pagestyle{empty}

\begin{abstract}
The proliferation of AI-generated images produced by Latent Diffusion Models (LDMs) has raised critical concerns regarding copyright infringement and misinformation. Although existing frequency-domain watermarking methods embed handcrafted geometric patterns into the initial latent noise prior to generation, they suffer from limited capacity and rigid pattern designs. We propose \textbf{DeepFreqMark}, an end-to-end learnable frequency-domain watermarking framework that replaces manual pattern engineering with a neural message encoder and decoder. To circumvent the computational bottleneck caused by Denoising Diffusion Implicit Model (DDIM) inversion during training, we introduce a Spherical Linear Interpolation (Slerp)-based attack simulation. This approach operates directly on the noise latent while strictly preserving the Gaussian variance. Extensive experiments demonstrate that DeepFreqMark achieves significantly lower Bit Error Rates (BER) than baseline methods under real-world attacks and scales to 256 bits message capacity. Our source code is available at \url{https://github.com/chenhsiu48/DeepFreqMark}.
\end{abstract}


\section{Introduction}

The rapid advancement of Latent Diffusion Models (LDMs) \cite{rombach2021highresolution}, such as Stable Diffusion \cite{podell2023sdxl}, has democratized high-quality image synthesis. While beneficial to creative industries, the proliferation of AI-generated content raises critical concerns about copyright infringement and misinformation. Consequently, developing robust, invisible watermarking techniques to trace the provenance of generated images is paramount.

Hur et al. \cite{hur2024latent} categorize watermarking methods for diffusion models into two paradigms: ``post-generation'' and ``in-generation.'' Post-generation methods apply watermarks to the final generated image. However, they are vulnerable to bypass attacks, and optimizing them for high robustness often severely degrades visual quality. In contrast, in-generation methods embed watermarks directly into the generative process, offering enhanced resilience while preserving image quality. 

Recent advances in LDM watermarking have shifted toward in-generation approaches. Methods such as Tree-Rings \cite{wen2023tree}, METR \cite{varlamov2024metr}, RingID \cite{ci2024ringid}, and HSTR/HSQR \cite{lee2025semantic} embed watermarks in the initial Gaussian-noise latent prior to the reverse diffusion process. Because the Denoising Diffusion Implicit Models (DDIMs) \cite{song2021denoising} inversion provides a deterministic mapping from the generated image back to the initial noise, these frequency-domain modifications appear as robust, invisible watermarks in the final pixel space. 

Despite their success, existing frequency-domain watermarking techniques suffer from two critical limitations. First, they rely on handcrafted geometric patterns embedded exclusively in the Fast Fourier Transform (FFT) domain. This rigid design limits the maximum message capacity and prevents generalization to other domains, such as the Discrete Cosine Transform (DCT). Second, existing methods assume that mid-to-low-frequency modifications naturally survive image-level attacks. They do not optimize for these attacks during the embedding phase because of the computational cost of DDIM inversions.

We introduce DeepFreqMark to address these challenges. Our contributions are summarized as follows:

\begin{itemize}
    \item \textbf{End-to-end Neural Embedding:} We propose a learnable framework that replaces rigid, handcrafted frequency patterns, extending semantic watermarking to both the DCT and FFT domains and scaling the payload capacity to 256 bits.
    \item \textbf{Slerp-based Attack Simulation:} We introduce a training mechanism that utilizes Spherical Linear Interpolation (Slerp) without DDIM inversion. By preserving the Gaussian variance of the noise latent, this simulation accurately approximates severe image-space distortions while avoiding the overhead of DDIM inversion.
    \item \textbf{Enhanced Robustness:} Extensive experiments demonstrate that DeepFreqMark achieves significantly lower Bit Error Rates (BER) than baseline methods across various real-world attacks, while preserving the visual fidelity and diversity of the generated images.
\end{itemize}

\section{Related Work}

Most initial-noise LDM watermarking methods use the FFT domain for embedding. For instance, Wen et al. \cite{wen2023tree} proposed Tree-Rings, which embeds concentric ring patterns in the low-frequency regions of the FFT spectrum. However, its limited capacity restricts its utility for multi-user tracking. To address this, Ci et al. \cite{ci2024ringid} introduced RingID to enhance multi-key identification capabilities. Similarly, METR \cite{varlamov2024metr} increased the message capacity by assigning binary bits to individual rings, enabling the encoding of up to $2^{16}$ messages. 

Despite these advancements, prior methods that used asymmetric frequency modifications discarded the imaginary parts during the inverse FFT, thereby compromising frequency-domain integrity. To mitigate this, HSTR/HSQR \cite{lee2025semantic} introduced Hermitian symmetric patterns to enforce real-valued constraints. 


\section{Methodology}

\subsection{Neural Frequency Embedding}

Our framework consists of a message encoder, $\mathcal{E}$, and a corresponding message decoder, $\mathcal{D}$. Given a binary message $m \in \{0, 1\}^L$, where $L$ denotes the message length in bits, the encoder $\mathcal{E}$ projects $m$ into a continuous space of size $C \times D_{\text{wm}} \times D_{\text{wm}}$, where $C$ is the channel dimension and $D_{\text{wm}} = 32$ is the spatial watermark dimension.

\begin{figure}[ht!]
\centering
\includegraphics[width=\columnwidth]{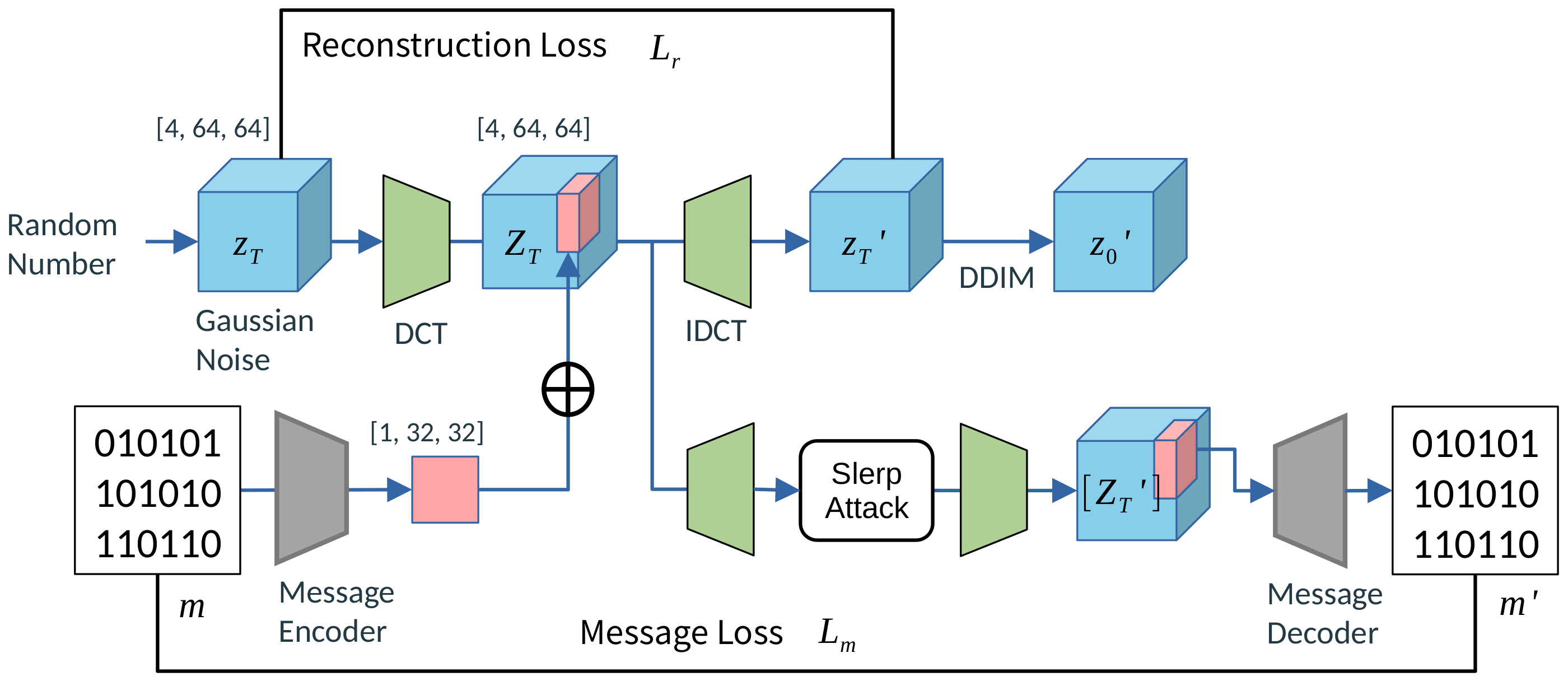}
\caption{DeepFreqMark embedding process in the DCT domain.}
\label{fig:dfm-dct}
\end{figure}

For the DCT Domain scheme in Fig. \ref{fig:dfm-dct}, the encoder outputs a single-channel, real-valued watermark $\Delta_{\text{DCT}}$, which is injected directly into the DCT-transformed latent via:

\begin{equation}
Z_T'=Z_T \oplus \Delta_{\text{DCT}},
\end{equation}

where $Z_T=\text{DCT}(z_T)$ is the DCT-transformed initial noise latent, $Z_T'$ denotes the watermarked noise latent in the frequency domain, and $\oplus$ signifies the element-wise addition applied to the upper-left, low-frequency region. This specific spectral band is selected to maximize robustness against common image-level attacks while minimizing distortion. 

\begin{figure}[ht!]
\centering
\includegraphics[width=\columnwidth]{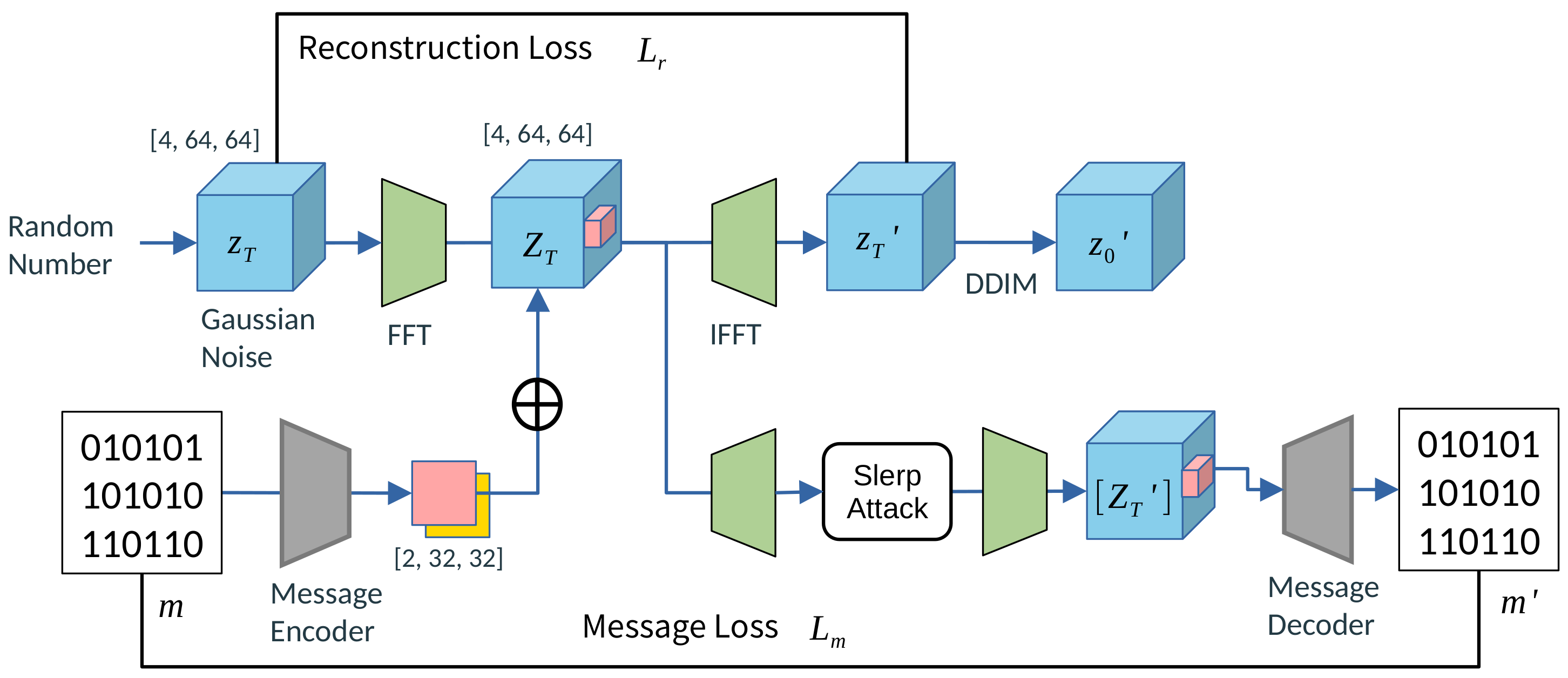}
\caption{DeepFreqMark embedding process in the FFT domain.}
\label{fig:dfm-fft}
\end{figure}

For the FFT Domain scheme in Fig. \ref{fig:dfm-fft}, the encoder outputs a two-channel, complex-valued watermark $\Delta_{\text{FFT}}$. To satisfy the Hermitian Symmetry requirement for generating real-valued spatial latents, we designate a ``free half-region'' located strictly to the right of the vertical DC axis. We then mathematically reflect and complex-conjugate this region to construct its symmetric counterpart:

\begin{equation}
F[M-k, N-l] = \overline{F[k, l]},
\end{equation}

thereby guaranteeing that the inverse FFT yields a real-valued noise latent. During this process, the FFT-transformed noise latent is center-shifted, and the complex watermark is injected into the central frequency region. In both the DCT and FFT domains, the watermark is embedded in the last channel of the transformed noise latent $Z_T$.

\subsection{Training Objective}

To train the message encoder and decoder end-to-end, we optimize a joint loss function that balances watermark imperceptibility in the latent space with message-extraction accuracy.

\textbf{Reconstruction Loss:} To ensure that the watermarked latent $z_T'$ remains statistically consistent with the initial Gaussian-noise latent $z_T \sim \mathcal{N}(0, \mathbf{I})$, we apply a Mean Squared Error (MSE) loss:

\begin{equation}
\mathcal{L}_r = \text{MSE}(z_T, z_T').
\end{equation}

Minimizing $\mathcal{L}_r$ guarantees that the frequency perturbation does not disrupt the variance or the generative stability of the downstream diffusion model.

\textbf{Message Loss:} To evaluate the accuracy of the extracted watermark, we employ the Binary Cross-Entropy (BCE) loss. The loss is computed between the original message $m$ and the recovered message $m'$ predicted by the decoder:

\begin{equation}
\mathcal{L}_m = \text{BCE}(m, m').
\end{equation}

The overall optimization objective is defined as a weighted sum of the reconstruction and message-extraction losses:

\begin{equation}
\mathcal{L} = \mathcal{L}_r + \alpha \mathcal{L}_m,
\end{equation}

where $\alpha$ is a balancing hyperparameter that governs the trade-off between preserving latent fidelity and maximizing message-retrieval accuracy.

\subsection{Watermark Encoder and Decoder Architecture}

\begin{figure}[ht!]
\centering
\includegraphics[width=\columnwidth]{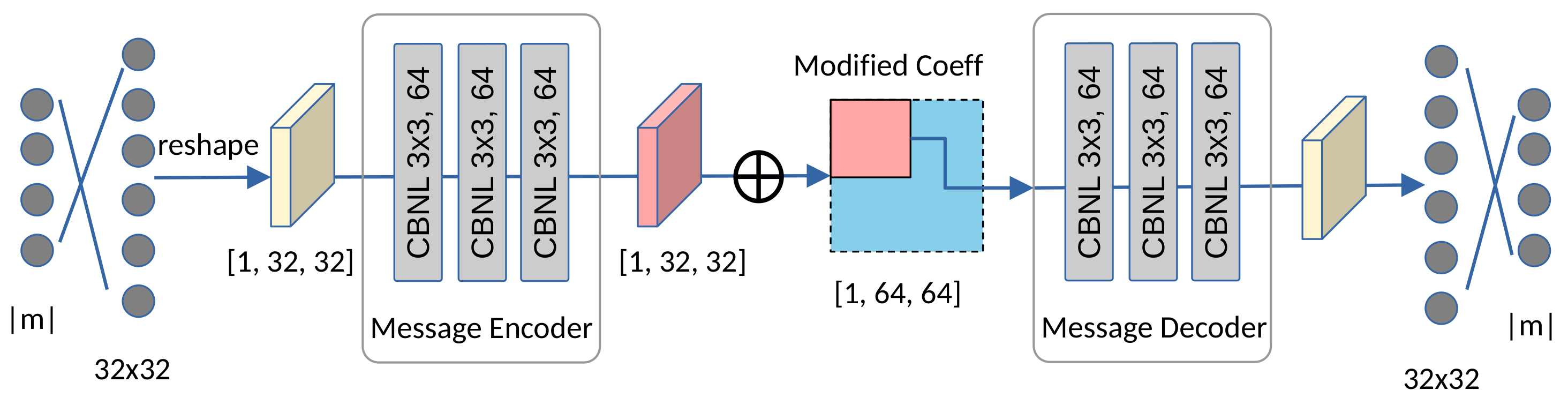}
\caption{Architectural overview of the message encoder and decoder for DCT domain embedding.}
\label{fig:dct-encdec}
\end{figure}

\begin{figure}[ht!]
\centering
\includegraphics[width=\columnwidth]{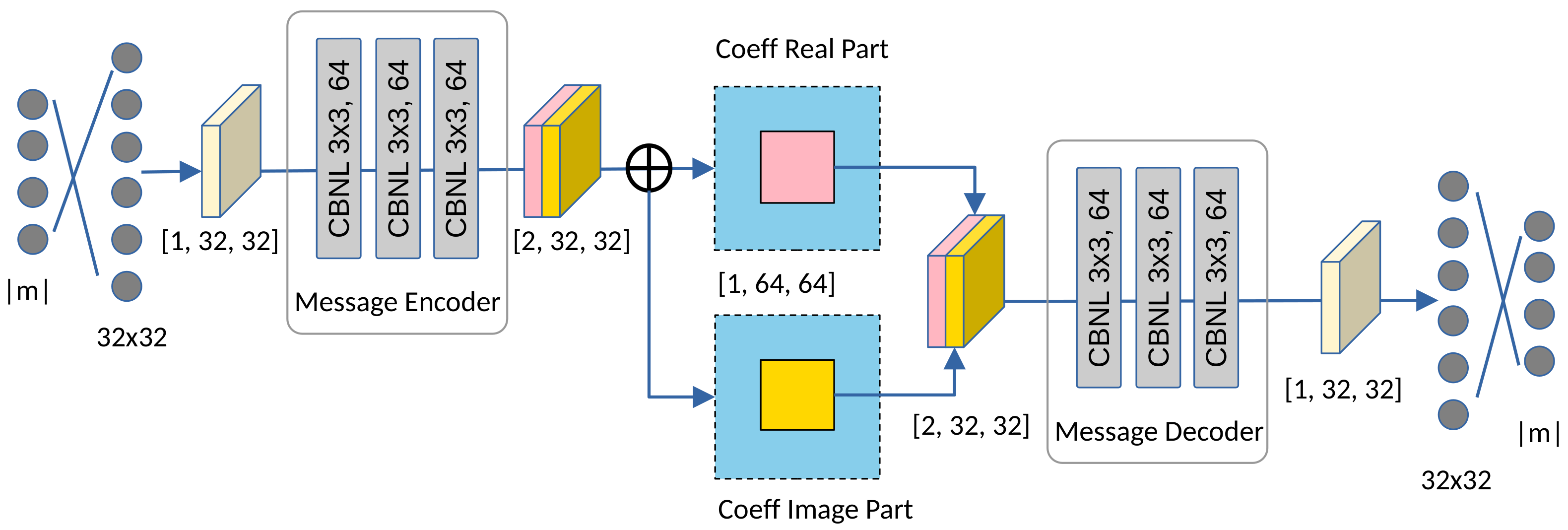}
\caption{Architectural overview of the message encoder and decoder for FFT domain embedding, utilizing a two-channel structure to separate real and imaginary components.}
\label{fig:fft-encdec}
\end{figure}

The message encoder and decoder networks are designed to learn an end-to-end continuous representation of a discrete binary message and its corresponding extraction logic within a localized $D_{\text{wm}} \times D_{\text{wm}}$ frequency patch. The detailed architectures of the encoder and decoder for both the DCT and FFT domains are illustrated in Figures \ref{fig:dct-encdec} and \ref{fig:fft-encdec}, respectively.

\textbf{Message Encoder ($\mathcal{E}$):} Given an input binary message $m \in \{0, 1\}^L$, the encoder first utilizes a fully connected linear layer to project the $L$-dimensional vector into a higher-dimensional representation. This intermediate continuous vector is then reshaped into a two-dimensional feature map of size $1\times 32\times 32$.

The reshaped tensor is then processed through a sequence of convolutional blocks. Each standard block comprises a $3 \times 3$ 2D convolutional layer with a stride of 1 and padding of 1, followed by 2D Batch Normalization (BN) to stabilize training and a LeakyReLU activation to introduce non-linearity. The final convolutional block omits the activation layer to allow an unconstrained continuous output range.

For the DCT framework, the final layer produces a real-valued watermark tensor $\Delta_{\text{DCT}} \in \mathbb{R}^{1 \times D_{\text{wm}} \times D_{\text{wm}}}$. Conversely, for the FFT framework, the output is a two-channel tensor $\Delta_{\text{FFT}} \in \mathbb{R}^{2 \times D_{\text{wm}} \times D_{\text{wm}}}$, representing the real and imaginary components of the complex watermark.

\textbf{Message Decoder ($\mathcal{D}$):} Let $I_w$ denote the generated watermarked image and $\tilde{I}_w$ its attacked counterpart. The recovered noise latent, denoted as $[z_T]$, is obtained via DDIM inversion:

\begin{equation}
[z_T] = \text{DDIM}^{-1}(\tilde{I}_w).
\end{equation}

During the message extraction phase, a localized frequency patch $\hat{\Delta}$ is isolated from the transformed attacked latent $[Z_T]$. For the DCT case, this is a single-channel patch $\hat{\Delta}_{\text{DCT}} \in \mathbb{R}^{1 \times D_{\text{wm}} \times D_{\text{wm}}}$ extracted from the upper-left, low-frequency region:

\begin{align}
[Z_T] &= \text{DCT}([z_T]), \\
\hat{\Delta}_{\text{DCT}} &= \ominus([Z_T]),
\end{align}

where $\ominus(\cdot)$ denotes the frequency coefficient cropping operator. For the FFT case, the $\ominus$ operator isolates the designated centered-frequency region and separates the complex values into real and imaginary channels, forming a two-channel input $\hat{\Delta}_{\text{FFT}} \in \mathbb{R}^{2 \times D_{\text{wm}} \times D_{\text{wm}}}$.

The message decoder $\mathcal{D}$ mirrors the encoder's structure with a symmetric series of convolutional blocks to distill watermark features. The resulting feature map is flattened into a 1D vector and passed through a fully connected layer that maps the features back to the original message dimensionality $L$, thereby recovering the binary sequence:

\begin{equation}
m' = \mathcal{D}(\hat{\Delta}).
\end{equation}

\subsection{Spherical Attack Simulation (Slerp)}

A core challenge in training robust watermarking frameworks for LDMs is the computational bottleneck of simulating real-world image distortions. Empirically, when standard image-level attacks (such as JPEG compression or Gaussian blur) are applied to a generated image, the subsequent DDIM inversion yields an attacked noise latent $[z_T]$ that exhibits a significant Mean Squared Error (MSE) relative to the original watermarked latent $z_T$, as reported in Table \ref{tab:mse_attacks}. In this context, the cheng2020-anchor\_3 and bmshj2018-factorized\_3 attacks represent neural image re-compression \cite{cheng2020learned} \cite{balle2018variational} at quality level 3. Concurrently, the diff\_attacker\_60 attack employs a diffusion-based purification model \cite{zhao2024invisible} designed to remove the watermark.

\begin{table}[!ht]
\centering
\caption{MSE of DDIM inversion of noise latents under various image attacks.}
\label{tab:mse_attacks}
\resizebox{\columnwidth}{!}{%
\begin{tabular}{@{}lc p{0cm} lc@{}}
\toprule
\textbf{Attack Type} & \textbf{MSE Error} & & \textbf{Attack Type} & \textbf{MSE Error} \\ \midrule
cheng2020-anchor\_3 & 0.8077 & & brightness\_0.5 & 0.1007 \\
bmshj2018-factorized\_3 & 0.8123 & & contrast\_0.5 & 0.1230 \\
diff\_attacker\_60 & 0.6286 & & Gaussian\_noise & 0.9091 \\
jpeg\_attacker\_50 & 0.6010 & & Gaussian\_blur & 0.4024 \\ \bottomrule
\end{tabular}
}
\end{table}

However, computing the full DDIM inversion loop at every training iteration to optimize for robustness is computationally prohibitive. To bridge this gap efficiently, DeepFreqMark introduces a surrogate attack simulation that operates directly on the noise latent. According to the Gaussian Annulus Theorem, the probability mass of a high-dimensional Gaussian distribution $\mathcal{N}(0, \mathbf{I})$ is heavily concentrated within a narrow spherical shell, effectively forming a high-dimensional ``hollow soap bubble.'' Consequently, simulating latent distortions via standard linear interpolation (Lerp) is mathematically flawed; it cuts through the interior of the hypersphere, artificially shrinking the variance and destroying the generative stability of the LDM. To resolve this, we employ Spherical Linear Interpolation (Slerp) to accurately traverse the surface of this hypersphere, simulating the trajectory of an attack via:

\begin{equation}
\text{Slerp}(z_T', z_r, s) = \frac{\sin((1-s)\Omega)}{\sin(\Omega)} z_T' + \frac{\sin(s\Omega)}{\sin(\Omega)} z_r,
\end{equation}

where $z_T'$ represents the watermarked noise latent, and $z_r \sim \mathcal{N}(0, \mathbf{I})$ denotes a random Gaussian noise vector that serves as the target direction for the destructive attack energy. The scalar parameter $\Omega$ denotes the angular distance between $z_T'$ and $z_r$. To simulate varying degrees of distortion, the attack strength parameter $s$ is uniformly sampled during training via $s \sim \mathcal{U}(0, S)$, where $S$ defines the upper bound of the allowable attack strength.

Because Slerp operates strictly on the spherical manifold, it is mathematically guaranteed to preserve the unit variance and the underlying Gaussian profile of the noise. This surrogate allows us to explicitly optimize the encoder and decoder to withstand severe image-space distortions by proxying attacks directly on the spatial noise latent, thereby bypassing the computationally expensive overhead of DDIM inversion.

\section{Experiments}

\subsection{Experimental Setup}

We train separate DeepFreqMark models for both the DCT and FFT embedding domains, each with a message capacity of 32 bits. To train the encoder and decoder networks, we randomly sample initial noise latents $z_T \sim \mathcal{N}(0, \mathbf{I})$ in batches of 100. The training procedure spans 300 epochs, with each epoch consisting of 5,000 steps. For each latent, we generate a random binary message and apply the corresponding frequency-domain embedding. Throughout training, we apply a Slerp-based attack simulation with a maximum strength of $S=0.6$. 

To evaluate the robustness of the trained models, we test DeepFreqMark with Stable Diffusion v1.5 \cite{rombach2021highresolution}. We randomly sample 100 text prompts from the DiffusionDB dataset \cite{wangDiffusionDBLargescalePrompt2022}. For each prompt, we generate a watermarked image using the trained encoder, then apply a suite of common image attacks, as listed in Table \ref{tab:mse_attacks}, to simulate real-world distortions. Finally, we perform DDIM inversion on the attacked images to recover the noise latent and extract the message to compute the Bit Error Rate (BER).


\subsection{Robustness}

Table \ref{tab:res_attack} presents the BER of the extracted messages under distinct attacks for both the DCT and FFT embedding methods, comparing models trained without ($S=0$) and with ($S=0.6$) the Slerp-based attack simulation. The empirical results indicate that the Slerp simulation substantially improves robustness, yielding a significant reduction in average BER across all attacks.

\begin{table}[ht!]
\centering
\caption{Comparison of BER across DCT and FFT Methods}
\label{tab:res_attack}
\resizebox{\columnwidth}{!}{%
\begin{tabular}{@{}lcccc@{}}
\toprule
& \textbf{DCT, S=0} & \textbf{DCT, S=0.6} & \textbf{FFT, S=0} & \textbf{FFT, S=0.6} \\ \midrule
Un-attacked           & 0.1250\% & 0.0000\% & 0.0938\% & 0.0000\% \\
cheng2020-anchor\_3   & 2.2500\% & 0.9375\% & 3.9063\% & 0.8750\% \\
bmshj2018-factorized\_3 & 1.8750\% & 0.7813\% & 3.6875\% & 0.5625\% \\
diff\_attacker\_60     & 2.4688\% & 1.0000\% & 3.0938\% & 0.3750\% \\
jpeg\_attacker\_50     & 0.6250\% & 0.3438\% & 1.0000\% & 0.0313\% \\
brightness\_0.5        & 0.1875\% & 0.0625\% & 0.0625\% & 0.0000\% \\
contrast\_0.5          & 0.1563\% & 0.1250\% & 0.1250\% & 0.0000\% \\
Gaussian\_noise        & 2.0313\% & 0.9375\% & 5.3438\% & 0.9688\% \\
Gaussian\_blur         & 0.2500\% & 0.0625\% & 0.4688\% & 0.0313\% \\ \midrule
\textbf{Average}      & \textbf{1.1076\%} & \textbf{0.4722\%} & \textbf{1.9757\%} & \textbf{0.3160\%} \\ \bottomrule
\end{tabular}
}
\end{table}

Table \ref{tab:method_comparison} compares DeepFreqMark with existing handcrafted, FFT-based watermarking methods in terms of message reconstruction accuracy and payload capacity, using performance metrics reported directly in the original publications. Unlike these baseline methods, which rely on a two-stage ``verification-then-identification'' pipeline, DeepFreqMark directly decodes the precise message bits from the transformed noise latents. Crucially, our learnable framework achieves significantly higher bit-extraction accuracy and a greater embedding capacity.

\begin{table}[!ht]
\centering
\caption{Comparison of Accuracy and Capacity across different watermarking methods.}
\label{tab:method_comparison}
\begin{tabular}{@{}lrr@{}}
\toprule
\textbf{Method} & \textbf{Accuracy} & \textbf{Capacity} \\ \midrule
Ours (FFT, S=0.6) & \textbf{99.68\%} & 32 bits \\
Ours (FFT, S=0.6) & \textbf{99.44\%} & 256 bits \\
Tree-Rings \cite{wen2023tree}$^\ast$ & 99.50\% & 0 bits \\
Tree-Rings \cite{wen2023tree} & 7.70\% & 11 bits \\
RingID \cite{ci2024ringid} & 94.20\% & 11 bits \\
METR \cite{varlamov2024metr} & 85.45\% & 16 bits \\
HSTR \cite{lee2025semantic}$^\ast$ & 88.90\% & 0 bits \\
HSQR \cite{lee2025semantic} & 98.50\% & 72 bits \\ \bottomrule
$^\ast$ Verification accuracy.
\end{tabular}
\end{table}

\subsection{Qualitative Results}

\begin{figure}[ht!]
\centering
\includegraphics[width=\columnwidth]{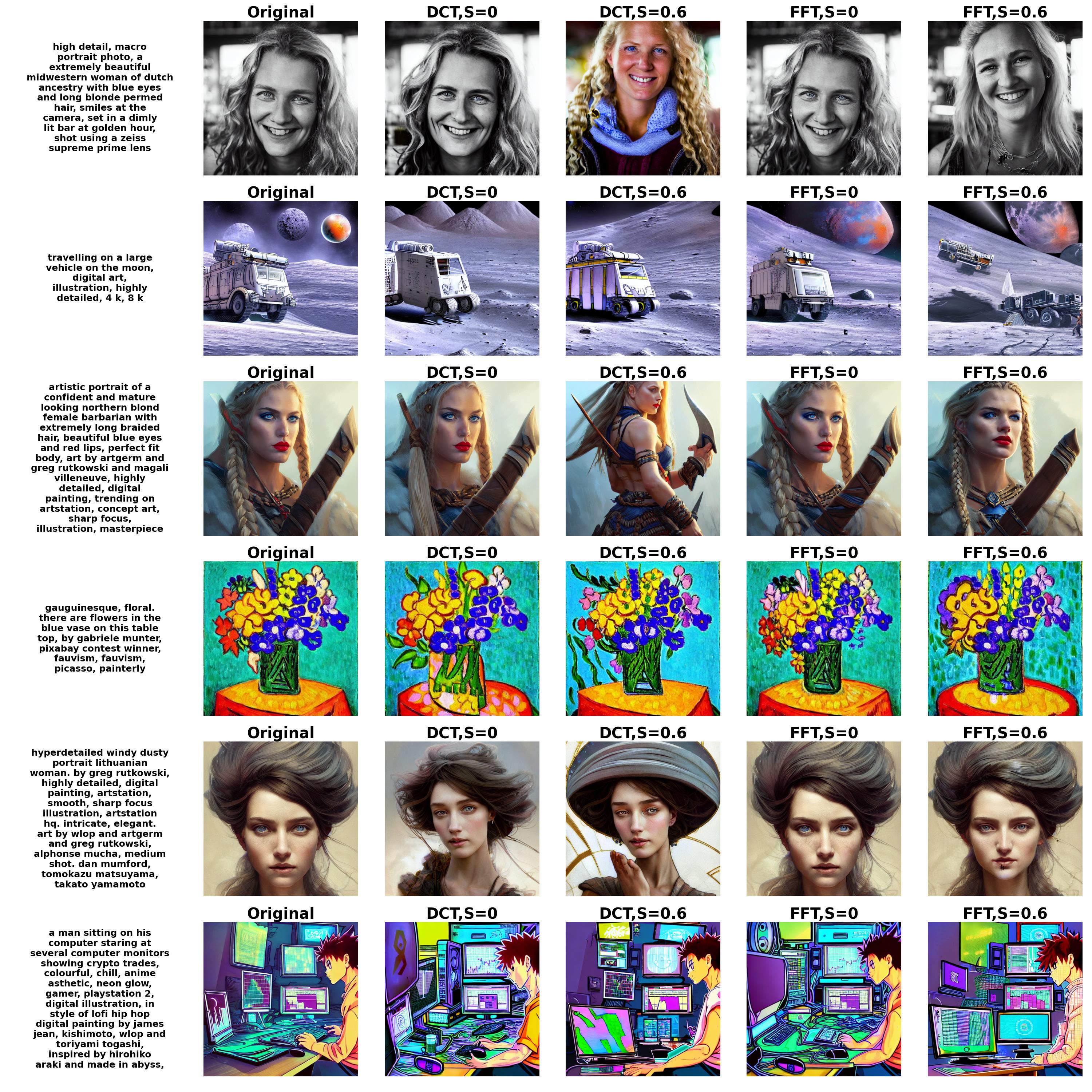}
\caption{Qualitative comparison of images generated via DeepFreqMark. The first column lists the input prompts, followed by the original unwatermarked images. The subsequent columns display watermarked images generated with $S=0$ and $S=0.6$ for both the DCT and FFT models.}
\label{fig:qual_attack}
\end{figure}

The qualitative results in Fig. \ref{fig:qual_attack} further illustrate the visual impact of training with and without the Slerp attack simulation. The figure provides a side-by-side comparison of the original unwatermarked images and the watermarked outputs from the DCT and FFT models. As observed, models trained without attack simulation ($S=0$) produce images that are slightly more coherent with the unwatermarked baseline, but their embedded signals are more vulnerable to attacks. Empirically, setting the attack strength to $S=0.6$ yields images that maintain high semantic fidelity to the original prompt while improving watermark resilience. 

\begin{figure}[ht!]
\centering
\includegraphics[width=\columnwidth]{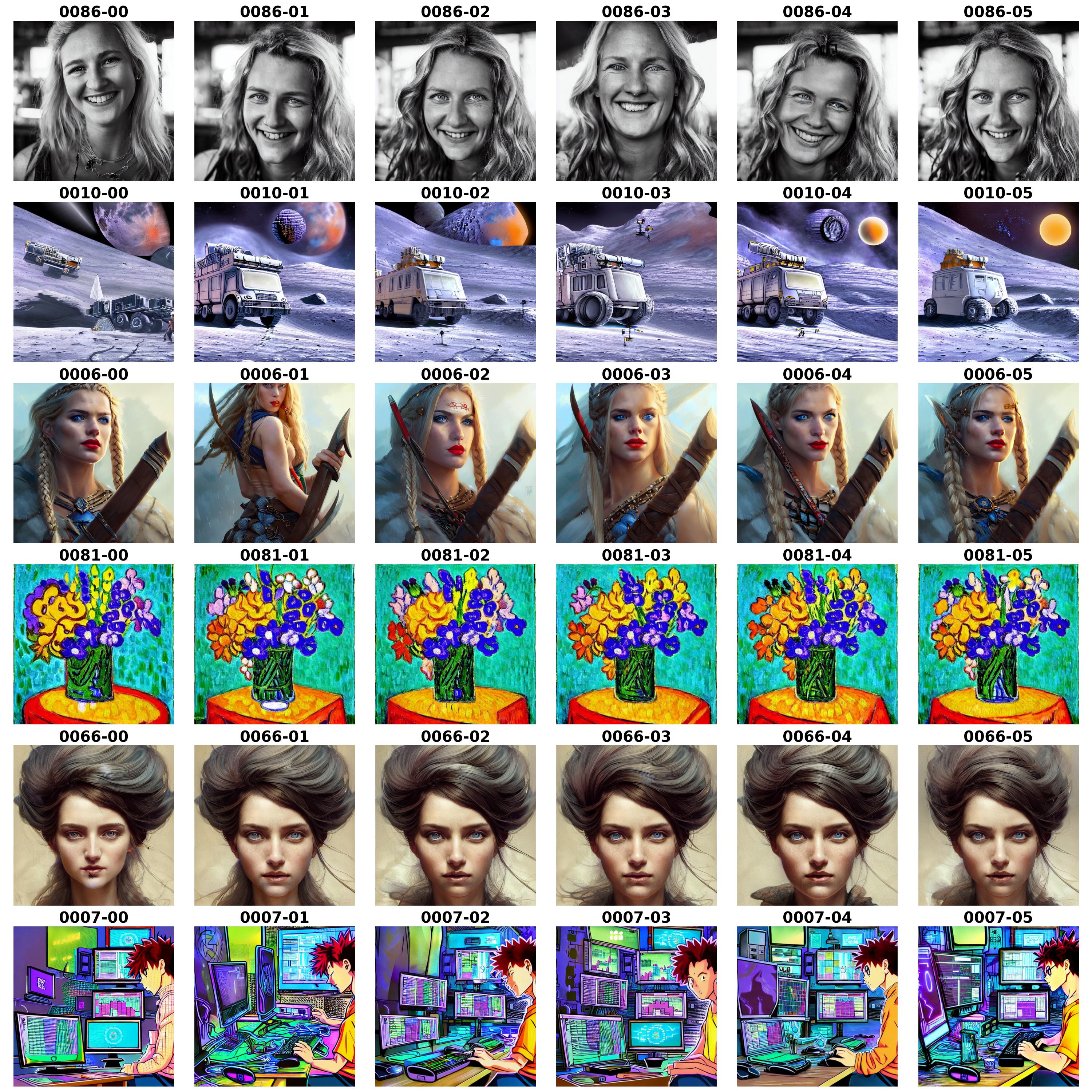}
\caption{Visual diversity of images generated using DeepFreqMark (FFT, $S=0.6$) for the same prompt across different watermark messages.}
\label{fig:qual_variety}
\end{figure}

Furthermore, Fig. \ref{fig:qual_variety} illustrates the visual diversity of images generated from the same prompt under different watermark messages. This diversity stems from the learnable nature of our neural message encoder, which optimizes spectral modifications to balance imperceptibility and robustness. Despite variations in the underlying messages, the generated images remain highly coherent with the target prompt. This confirms that the framework does not converge to a rigid, fixed embedding pattern, thereby enabling a vast and unique message space for provenance tracking.

\subsection{Attack Strength}

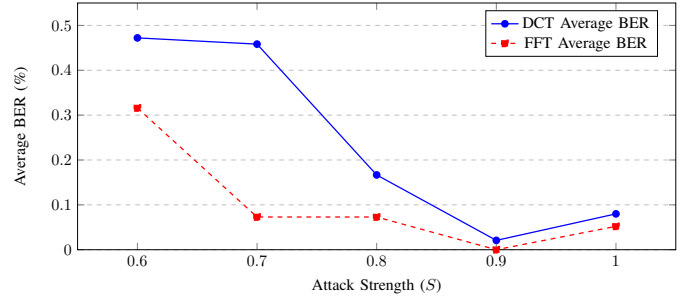
\begin{figure}[!ht]
    \centering
\resizebox{\columnwidth}{!}{%
    \begin{tikzpicture}
    \begin{axis}[
        xlabel={Attack Strength ($S$)},
        ylabel={Average BER (\%)},
        xmin=0.55, xmax=1.05,
        ymin=0, ymax=0.55,
        xtick={0.6, 0.7, 0.8, 0.9, 1.0},
        ytick={0, 0.1, 0.2, 0.3, 0.4, 0.5},
        legend pos=north east,
        ymajorgrids=true,
        grid style=dashed,
        width=0.8\textwidth,
        height=7cm,
        tick label style={/pgf/number format/fixed}
    ]

    \addplot[
        color=blue,
        mark=*,
        thick,
    ]
    coordinates {
        (0.6, 0.4722) (0.7, 0.4583) (0.8, 0.1667) (0.9, 0.0208) (1.0, 0.0799)
    };
    \addlegendentry{DCT Average BER}

    \addplot[
        color=red,
        mark=square*,
        dashed,
        thick,
    ]
    coordinates {
        (0.6, 0.3160) (0.7, 0.0729) (0.8, 0.0729) (0.9, 0.0000) (1.0, 0.0521)
    };
    \addlegendentry{FFT Average BER}

    \end{axis}
    \end{tikzpicture}
}
    \caption{Average BER for DCT and FFT across attack strength $S$.} \label{fig:strength_scaling}
\end{figure}

To investigate the impact of attack strength in the Slerp simulation, we evaluate the BER across a range of maximum attack strengths, $S \in \{0.6, 0.7, 0.8, 0.9, 1.0\}$. As illustrated in Fig. \ref{fig:strength_scaling}, increasing the maximum attack strength $S$ generally leads to a reduction in BER. Specifically, the configuration with $S=0.9$ achieves the lowest BER.

\begin{figure}[ht!]
\centering
\includegraphics[width=\columnwidth]{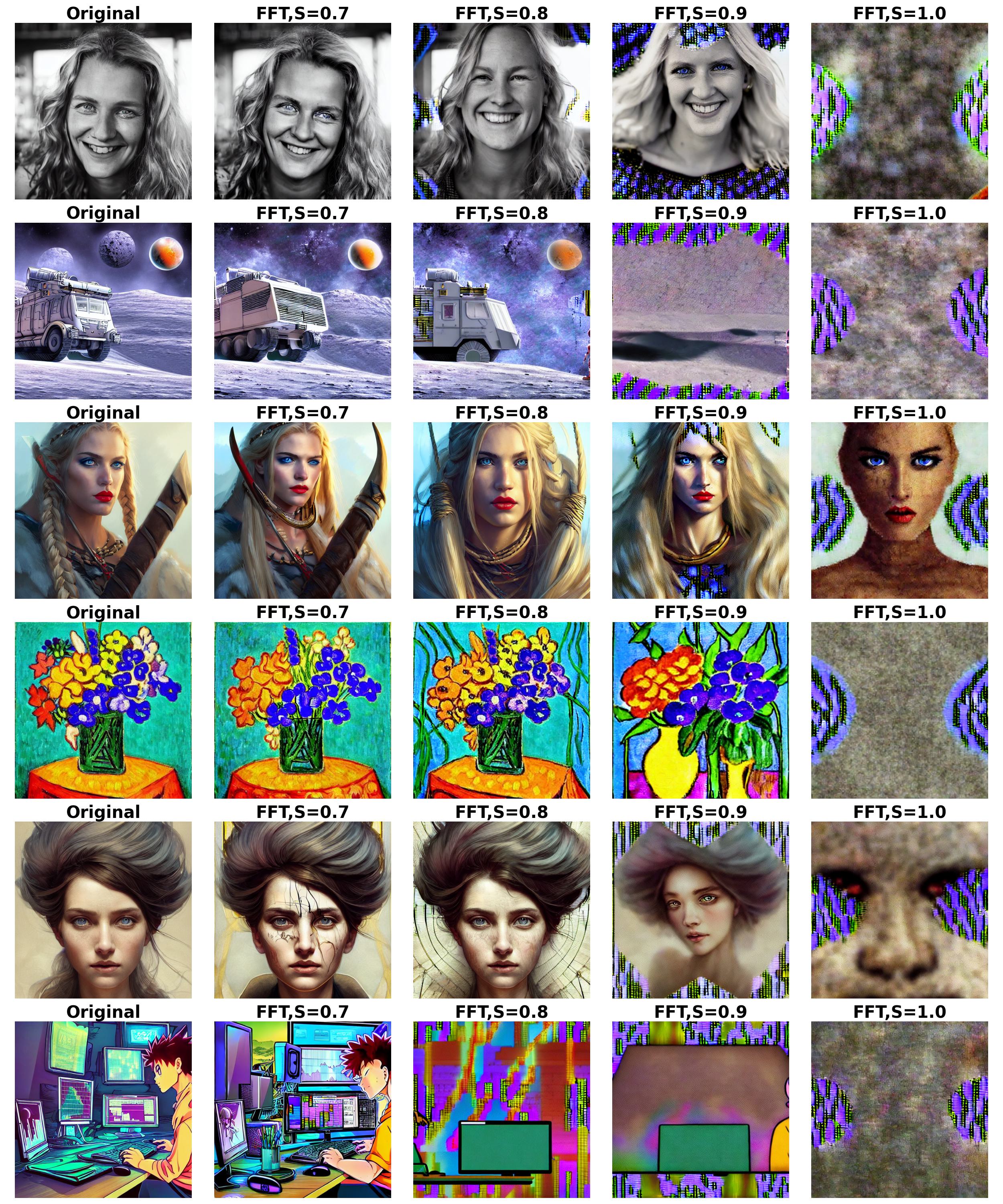}
\caption{Visual quality degradation of watermarked images under elevated attack strengths ($S \ge 0.8$).}
\label{fig:qual_cost}
\end{figure}

However, this improvement in robustness comes at a cost. As shown in Fig. \ref{fig:qual_cost}, the visual quality of the generated images degrades as $S$ increases, with noticeable distortion when $S \ge 0.8$. The images become increasingly distorted and less aligned with the original prompt, exhibiting noticeable artifacts and a loss of fine details. This perceptual degradation occurs because a higher attack strength forces the encoder to embed the watermark more aggressively to ensure its survival. While these aggressive spectral modifications improve robustness, they destabilize the diffusion model's generation trajectory, highlighting a clear trade-off between watermark resilience and visual fidelity.

\subsection{Message Length}

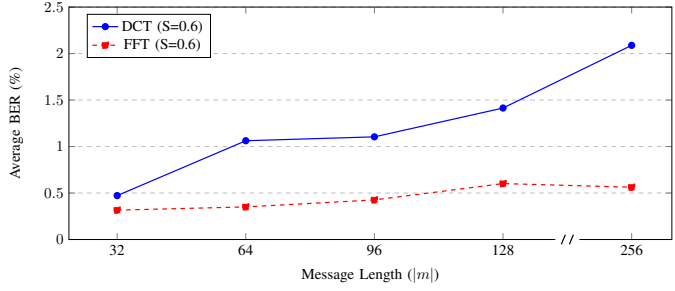
\begin{figure}[!ht]
    \centering
    \resizebox{\columnwidth}{!}{%
    \begin{tikzpicture}
    \begin{axis}[
        xlabel={Message Length ($|m|$)},
        ylabel={Average BER (\%)},
        xmin=20, xmax=170, 
        ymin=0, ymax=2.5,
        xtick={32, 64, 96, 128, 160}, 
        xticklabels={32, 64, 96, 128, 256}, 
        ytick={0, 0.5, 1.0, 1.5, 2.0, 2.5},
        legend pos=north west,
        ymajorgrids=true,
        grid style=dashed,
        width=0.85\textwidth,
        height=7cm,
        clip=false 
    ]

    \addplot[color=blue, mark=*, thick]
    coordinates {
        (32, 0.4722) (64, 1.0625) (96, 1.1042) (128, 1.4141) (160, 2.0885)
    };
    \addlegendentry{DCT (S=0.6)}

    \addplot[color=red, mark=square*, dashed, thick]
    coordinates {
        (32, 0.3160) (64, 0.3507) (96, 0.4259) (128, 0.6016) (160, 0.5625)
    };
    \addlegendentry{FFT (S=0.6)}


    \draw [white, line width=3pt] (121, 0) -- (127, 0);

    \draw [thick] (122.5, -5.15) -- (123.5, 5.15);
    \draw [thick] (124.5, -5.15) -- (125.5, 5.15);

    \end{axis}
    \end{tikzpicture}
    }
    \caption{Average BER vs. message length (at $S=0.6$). The gap between $|m|=128$ and $|m|=256$ is shortened for better visualization.}
    \label{fig:length_scaling}
\end{figure}

Since our proposed DeepFreqMark framework is designed to be flexible across different message lengths, we evaluate BER performance across a range of message lengths $|m| \in \{32, 64, 96, 128, 256\}$ bits. As illustrated in Fig. \ref{fig:length_scaling}, BER scales proportionally with the message length; the BER remains below 1.5\% for both the DCT and FFT models for payloads up to 128 bits. In practical scenarios, such low error rates can be seamlessly corrected using standard error-correcting codes, such as Bose-Chaudhuri-Hocquenghem (BCH) codes.

\subsection{Superiority of FFT over DCT}

Comparing the DCT and FFT embedding methods in Figures \ref{fig:strength_scaling} and \ref{fig:length_scaling}, we observe that the FFT-based DeepFreqMark variants consistently achieve a lower BER across varying attack strengths and message lengths. 

The superior robustness of the FFT-based approach stems from its complex-valued embedding space. Unlike real-valued DCT coefficients, FFT provides the encoder with magnitude and phase components, granting greater degrees of freedom. Furthermore, the mandatory Hermitian Symmetry required for real-valued latents introduces structural redundancy in the FFT spectrum, granting the decoder natural resilience against magnitude-distorting and quantization attacks.

\section{Conclusion}

We presented DeepFreqMark, a novel end-to-end neural frequency-domain watermarking framework designed to trace the provenance of images generated by LDMs. By replacing rigid, handcrafted geometric patterns with a learnable embedding strategy, our method successfully generalizes across both the DCT and FFT domains and scales the payload capacity to 256 bits. A core innovation of our approach is the Slerp-based spherical attack simulation, which preserves the Gaussian distribution of the noise latent while efficiently proxying real-world image distortions. Extensive experiments demonstrate the effectiveness and scalability of our framework. Consequently, DeepFreqMark provides a highly resilient, scalable, and visually coherent solution for generative content attribution. Our source code is available at \url{https://github.com/chenhsiu48/DeepFreqMark}. 

\section*{Acknowledgments}
The author gratefully acknowledges support from the National Science and Technology Council (NSTC) of Taiwan, which sponsored this joint research with Dr. Köppen during his visiting research stay at Kyushu Institute of Technology under Grant No. 114-2917-I-002-022.

\printbibliography

\end{document}